\documentclass{article}

\usepackage{PRIMEarxiv}

\usepackage[utf8]{inputenc} 
\usepackage[T1]{fontenc}    
\usepackage{hyperref}       
\usepackage{url}            
\usepackage{booktabs}       
\usepackage{amsfonts}       
\usepackage{nicefrac}       
\usepackage{microtype}      
\usepackage{lipsum}
\usepackage{fancyhdr}       
\usepackage{graphicx}       
\usepackage{multirow}
\usepackage{mathtools}

\usepackage[inkscapelatex=false]{svg}
\graphicspath{{media/}}     

\title{SGIA: Enhancing Fine-Grained Visual Classification with Sequence Generative Image Augmentation
}

\author{
  Qiyu Liao, Xin Yuan \\
  Data61, CSIRO  \\
  \texttt{\{qiyu.liao, xin.yuan\}@csiro.au} \\
   \And
  Min Xu \\
  University of Technology Sydney \\
  \texttt{min.xu@uts.edu.au} \\
   \And
  Dadong Wang \\
  Data61, CSIRO \\
  \texttt{dadong.wang@csiro.au} \\
}

\begin{document}
\maketitle

\begin{abstract}
	
In Fine-Grained Visual Classification (FGVC), distinguishing highly similar subcategories remains a formidable challenge, often necessitating datasets with extensive variability. The acquisition and annotation of such FGVC datasets are notably difficult and costly, demanding specialized knowledge to identify subtle distinctions among closely related categories. Our study introduces a novel approach employing the Sequence Latent Diffusion Model (SLDM) for augmenting FGVC datasets, called Sequence Generative Image Augmentation (SGIA). Our method features a unique Bridging Transfer Learning (BTL) process, designed to minimize the domain gap between real and synthetically augmented data. This approach notably surpasses existing methods in generating more realistic image samples, providing a diverse range of pose transformations that extend beyond the traditional rigid transformations and style changes in generative augmentation. We demonstrate the effectiveness of our augmented dataset with substantial improvements in FGVC tasks on various datasets, models, and training strategies, especially in few-shot learning scenarios. Our method outperforms conventional image augmentation techniques in benchmark tests on three FGVC datasets, showcasing superior realism, variability, and representational quality. Our work sets a new benchmark and outperforms the previous state-of-the-art models in classification accuracy by 0.5\% for the CUB-200-2011 dataset and advances the application of generative models in FGVC data augmentation.

\end{abstract}    
\section{Introduction}
\label{sec:intro}

In the rapidly evolving field of computer vision, Fine-Grained Visual Classification (FGVC) stands out as a discipline that delves into the minutiae of object distinctions within highly specialized categories. This precision-focused area of study, which has been the subject of increasing interest, requires identifying subtle differences among objects, such as various species of birds \cite{wah2011caltech} or intricate car models \cite{krause20133d}. Unlike general image classification that broadly categorizes images, FGVC challenges algorithms to discern between closely related categories, necessitating a depth of detail and variability that far exceeds that of conventional image classification datasets.

\begin{figure}[t!]
    \includesvg[width=\linewidth]{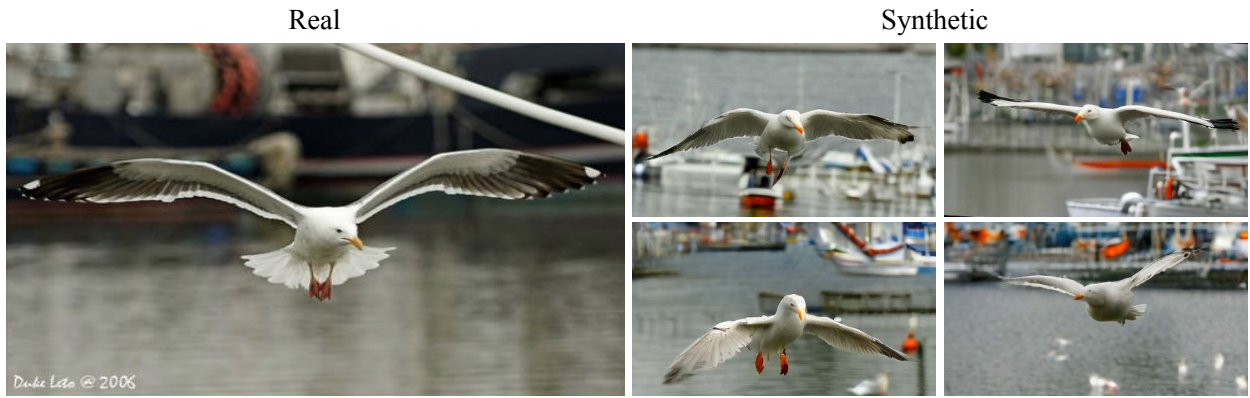}
	\caption{Illustration of synthetic image quality. The left image is from the CUB-2011-200 dataset. The four images on the right are synthetic ones generated from the original.}
	\label{COVER}
\end{figure}

Historically, enhancing the nuanced discriminative power of FGVC systems has been approached through various methodologies. Early efforts concentrated on expanding the feature space through higher-order feature expansion techniques \cite{lin2015bilinear, gao2016compact, rahman2023learning}, thereby enriching the representational depth of neural networks. Concurrently, there has been a surge in employing attention mechanisms \cite{zheng2017learning, ding2019selective, hu2019better}, aimed at isolating and emphasizing critical features of target objects. More recently, attention has shifted towards models that facilitate superior feature learning and detail localization through innovative attention-based frameworks \cite{rao2021counterfactual, liao2022category, bera2022sr}. Despite these advancements, the construction of comprehensive and diverse FGVC datasets remains a formidable challenge, exacerbated by cost, privacy, and copyright constraints. In this context, data augmentation emerges as a crucial strategy, not only mitigating these challenges by enriching dataset variability without additional data collection but also enhancing the robustness and generalization of FGVC models.

Conventional dataset augmentation strategies, such as rotations, flips, and color adjustments have been pivotal in enhancing the diversity of datasets \cite{yang2022image}. Despite their utility, these conventional methods often fail to introduce the level of variability required to meet the nuanced demands of FGVC tasks. The evolution of generative models, capable of mimicking real-data distributions, marks a significant advancement, enabling the creation of high-fidelity and photorealistic images. Particularly, breakthroughs in text-to-image generation models, as highlighted by \cite{ho2020denoising}, have emerged as notable achievements in generating high-quality images from textual descriptions, offering a novel avenue for image data augmentation. Furthermore, \cite{he2023is} delved into Generative Image Augmentation (GIA) across zero-shot, few-shot, and general image classification tasks. Yet, the approach of generating images from scratch, while preserving the original image structure, often results in limited style variations or minimal changes in texture or background. This not only marginally enhances representation but also introduces significant domain bias, adversely affecting FGVC performance, particularly with large datasets. The necessity for more sophisticated data augmentation methods is therefore underscored, emphasizing their crucial role in overcoming these challenges and propelling the field of FGVC forward.

Addressing the aforementioned challenges, this paper introduces a novel framework for image augmentation, centered around the Sequence Latent Diffusion Model (SLDM), designed to inject a wide array of variations into FGVC images. This approach marks a significant departure from traditional Generative Image Augmentation (GIA) methods, offering unparalleled diversity in texture, position, angle, motion, and background settings without sacrificing image quality or the integrity of discriminative features. The key contributions of our work can be summarized as follows:

\begin{itemize}
	\item We conducted a comprehensive exploration of the limitations inherent in applying diffusion models for fine-grained image augmentation. In doing so, we introduced the SLDM for Sequence Generative Image Augmentation (SGIA), showcasing its superiority in generating detailed and varied images that maintain high fidelity to the original data (as illustrated in Fig. \ref{COVER}).
	\item We introduced a novel Bridging Transfer Learning (BTL) strategy, designed to effectively close the gap between source datasets and their augmented counterparts. This methodology ensures that the enhanced datasets preserve a high degree of generalizability and accuracy, facilitating seamless application in FGVC tasks.
	\item Our study evaluated the SGIA and BTL methods across diverse datasets, models, augmentations, and image sizes, demonstrating notable accuracy improvements in FGVC tasks. These findings not only confirmed the robustness of our proposed methods but also provided a practical guide for optimizing FGVC training configurations with SGIA.
	\item Through rigorous testing and evaluation, we demonstrated that our SGIA framework significantly improved the generalization capabilities of FGVC models. Our approach sets a new benchmark for performance on the CUB-2011-200 dataset, establishing the first instance of a Generative Image Augmentation technique that outperforms pure real datasets in large-scale FGVC challenges.
\end{itemize}

By pushing the boundaries of what is possible with generative-based image augmentation for FGVC, our research not only addresses the immediate challenges of dataset diversity and representational fidelity but also lays the baseline for future explorations in the field.
\section{Related Work}

The research landscape relevant to our study encompasses two pivotal domains: Fine-Grained Visual Classification (FGVC) and Generative Image Augmentation (GIA). Both areas have witnessed significant advancements, shaping the methodologies and technologies applicable to enhancing FGVC through improved image augmentation techniques.

\subsection{Fine-Grained Visual Classification (FGVC)}

FGVC methodologies have seen considerable evolution, with developments concentrating on enhancing the precision of classification within highly similar object categories. This evolution can be categorized into three primary streams:

{\textit{Part-based Approaches:}} This line of research emphasizes the identification and analysis of specific object parts to improve recognition accuracy. Notably, the MA-CNN architecture \cite{zheng2017learning} represents a leap forward by integrating feature map clustering with part localization to enhance classification precision. Similarly, S3N \cite{ding2019selective} leverages local category-specific responses to refine feature representation, while WS-DAN \cite{hu2019better} employs attention-driven multi-inference strategies for isolating discriminative features. These approaches underscore the significance of focusing on detailed object parts for fine-grained classification.

{\textit{Higher-Order Feature Expansion:}} Techniques under this category aim to amplify the capacity of convolutional neural networks (CNNs) to represent complex visual patterns through enhanced feature spaces. Bilinear CNNs \cite{lin2015bilinear} and their derivatives introduce sophisticated mechanisms for expanding and normalizing the feature matrix, thereby improving the model's ability to capture intricate visual details. Efforts to manage the dimensionality and computational load of these expanded features, such as compact matrix estimation \cite{gao2016compact} and selective feature compression \cite{liao2019squeezed}, address critical scalability and efficiency challenges inherent in higher-order methods.

{\textit{Attention-based Models:}} Leveraging attention mechanisms constitutes a dynamic and increasingly influential research area within FGVC. Models like MAMC \cite{sun2018multi}, API-Net \cite{zhuang2020learning}, and more advanced structures incorporating Graph Convolutional Networks (GCNs) and Transformer architectures, like SR-GNN \cite{bera2022sr}, exemplify the push towards more nuanced feature learning and object detail capture. These approaches benefit from the ability to dynamically focus on relevant aspects of an image, enhancing the model's discriminative power.

Recently, with the development of Transformer\cite{vaswani2017attention} in the computer vision field, many improved Vision Transformer architectures have been proposed, such as FFVT\cite{wang2021feature}, SIM-Trans\cite{sun2022sim}, TransFG\cite{he2022transfg}, MetaFormer\cite{diao2022metaformer}, and AFTrans\cite{zhang2022free}, these methods utilize self-attention maps in transformer layers to enhance feature learning and locate object details.

\subsection{Generative Image Augmentation (GIA)}

GIA represents a frontier in addressing the intrinsic challenges of FGVC, especially concerning the generation of detailed and diverse synthetic datasets. Early GIA approaches \cite{dosovitskiy2015flownet, richter2016playing, peng2017visda} generated synthetic datasets using traditional pipelines, but faced limitations in realism and diversity. 

The introduction of advanced generative models like class-conditional GANs~\cite{besnier2020dataset} and StyleGAN~\cite{karras2019style} has markedly improved the quality and applicability of synthetic data for training purposes. These models facilitate the creation of highly realistic images that can significantly augment existing datasets, improving the performance of classifiers across various tasks, including FGVC.

Recent explorations into the manipulation of GAN latent spaces and the application of diffusion models for generating viewpoint and feature-specific augmentations have opened new avenues for dataset enhancement. Techniques such as \cite{jahanian2021generative, ho2020denoising, he2023is} demonstrate the potential of diffusion models to contribute to the training of more robust and accurate classifiers by providing a diverse array of training examples. However, challenges remain in fine-tuning these generative approaches to maintain a delicate balance between introducing variability and preserving the essential characteristics of the target classes, particularly in the context of large-scale FGVC datasets\cite{he2023is}.

Our research situates itself at the confluence of these developments, aiming to leverage the latest in generative modeling to surmount the current limitations faced by FGVC methodologies. By introducing a novel augmentation framework that synergizes with fine-grained classification requirements, we aspire to push the boundaries of what is achievable in this challenging yet critical domain.
\section{Methodology}

\begin{figure*}[t!]
	\centering
    \includesvg[width=\linewidth]{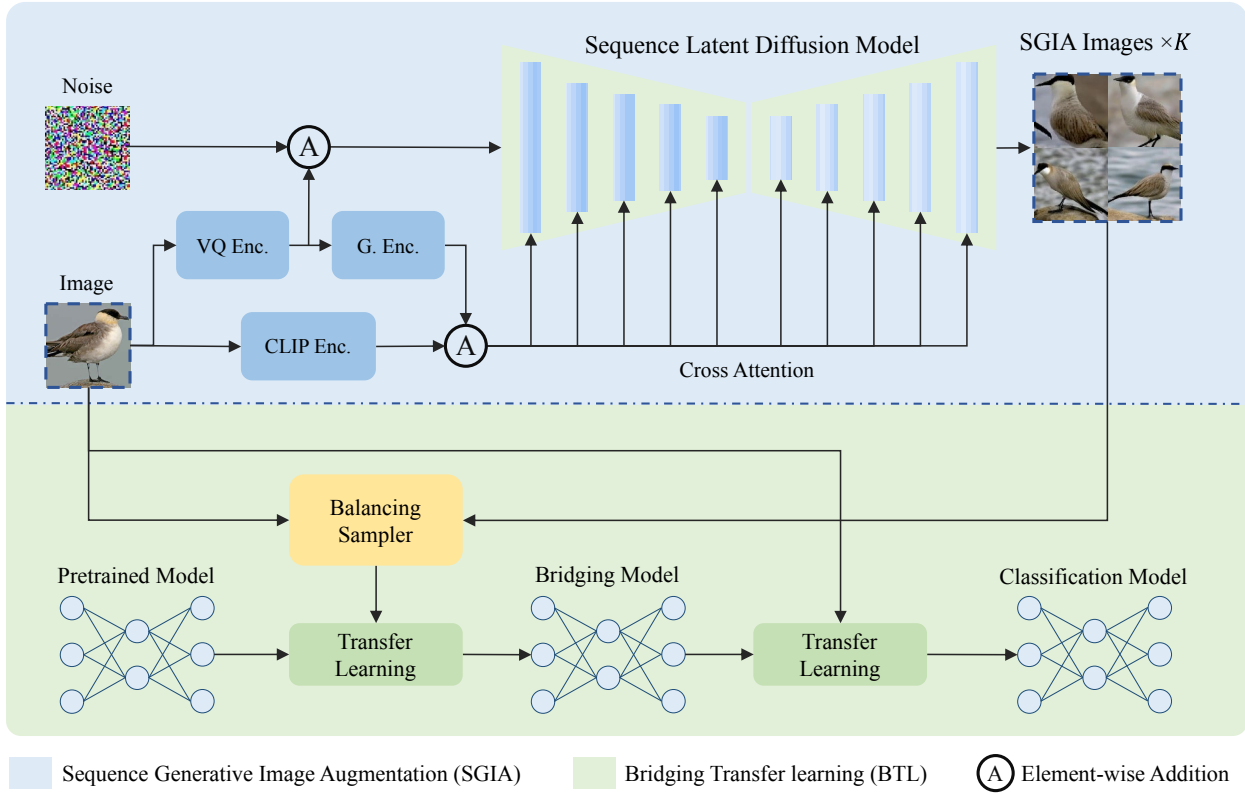}
	\caption{Two-phase neural network training framework. The process begins with encoding images with video motion and semantic features to guide the SLDM in the denoising phase. A Balancing Sampler then integrates augmented data with original data for the transfer learning of the bridging model. Finally, this model is fine-tuned on the original dataset to complete the classification model.}
	\label{FRAMEWORK}
\end{figure*}

Inspired by Generative Image Augmentation (GIA) approaches, which utilize GANs and diffusion models for creating synthetic image samples, we introduce the Sequence Generative Image Augmentation (SGIA) framework. Unlike image-based augmentation methods, SGIA leverages a sequence-based generator to infuse additional variations while maintaining the distinguishing characteristics of the primary object. Our approach integrates two main components: the Sequence Generative Image Augmentation (SGIA) and the Bridging Transfer Learning (BTL) process, as illustrated in Fig. \ref{FRAMEWORK}. Together, these mechanisms work in tandem to enrich FGVC datasets with the enhanced diversity and robustness required for accurate fine-grained classification.

\subsection{Sequence Generative Image Augmentation (SGIA)}

Drawing from the approach of VideoComposer\cite{wang2023videocomposer}, we employ the Latent Diffusion Model (LDM)\cite{rombach2022high} as the core model for our SGIA, herein referred to as Sequence-LDM (SLDM). We incorporate the frontend architecture of I2VGen-XL\cite{zhang2023i2vgen}, utilizing the image encoder from CLIP\cite{radford2021learning} to extract semantic features from images and employing the Encoder from VQGAN (VQ Enc.)\cite{esser2021taming} as the source of variations for our SGIA. Additionally, the Global Encoder (G. Encoder) from I2VGen-XL is used as a perceptor for the nuanced details of image categories. The outputs of the two ways of encoders are added and utilized to guide the SLDM in generating image sequences. Simultaneously, the output from the VQGAN Encoder is added to a random noise, which is then input into the SLDM for diffusion operations. By harnessing the VQGAN and SLDM's adeptness in motion and general knowledge perception, we produce image sequences from single images in the FGVC dataset. These sequences introduce variations in poses, angles, positions, and backgrounds while maintaining key characteristics of the original image.

In line with \cite{zhang2023i2vgen}, we adopt a variant of Latent Diffusion Models (LDMs) that operate in latent space, ensuring local fidelity and visual manifold preservation. We utilize the pretrained VQGAN Encoder $E_{VQ}$ from \cite{esser2021taming}, the pretrained Global Encoder $E_{G}$ from \cite{zhang2023i2vgen}, and the pre-trained CLIP image Encoder $E_{CLIP}$ from \cite{radford2021learning} for detail, global, and semantic feature extraction, respectively. For the input image $x$, we extract its CLIP image features and perform a simple addition operation with them on the output of $E_{VQ}$ and $E_{G}$, employing cross-attention to supervise each layer within the Semantic-Latent Diffusion Model (SLDM). Parallelly, the output from the VQGAN feature of the input image $E_{VQ}(x)$ is simply added to the noise $\epsilon\sim\mathcal{N}(0,I)$, which is then fed into the SLDM to generate $K$ augmented outputs:

\begin{equation}
	\tilde{x}={LDM}(\epsilon+E_{VQ}(x), E_{CLIP}(x)+E_{G}(E_{VQ}(x))),
\end{equation}

The resulting augmented sequence $\tilde{x}\in\mathbb{R}^{K\times H\times W\times3}$ is then utilized to generate mini-batches for subsequent model training stages.

\subsection{Balancing Real and Synthetic Data}

Training models exclusively on synthetic images can inadvertently emphasize spurious qualities and biases inherent in generative models. To mitigate this, a common practice is to assign different sampling proportions to real and synthetic images, as a means to manage potential imbalances \cite{he2023is}. We adopt a similar approach for balancing real and synthetic data, as detailed in Equation (\ref{balancingequa}), where $\alpha$ represents the probability of including a synthetic image at the $l$-th location in the training data loader $\mathcal{L}_{\alpha}(i)$:
\begin{equation}
	\begin{gathered}
		i \sim \mathcal{U}({1,\ldots,N}),\\
		j \sim \mathcal{U}({1,\ldots,M}),\\
		k \sim \mathcal{U}({1,\ldots,K}),\\
		\mathcal{L}_{\alpha}(i) \leftarrow X_i \text{ with probability } (1-\alpha) \text{ else } \tilde{X}_{ijk}
	\end{gathered}
	\label{balancingequa}
\end{equation}

In this framework, $X={x_1, x_2, \ldots, x_N} \in \mathcal{R}^{N \times H \times W \times 3}$ represents a dataset of $N$ real images. For each image $x_i$, we generate $M$ augmentation sequences, each containing $K$ synthetic images, yielding a synthetic dataset $\tilde{X} \in \mathcal{R}^{N \times M \times K \times H \times W \times 3}$ with $N \times M \times K$ image augmentations. The synthetic image ${\tilde{x}}_{ijk} \in \mathcal{R}^{H \times W \times 3}$ is the $k$-th image in the $j$-th sequence derived from the $i$-th real image. Indices $i$, $j$, and $k$ are randomly and uniformly sampled from the respective sets of $N$ real images, $M$ augmented sequences, and $K$ images within each sequence. Depending on the value of $\alpha$, a real image $x_i$ or its augmented counterpart $\tilde{x}_{ijk}$ is added to the loader $\mathcal{L}_{\alpha}(i)$. In line with \cite{zhang2023i2vgen}, we set the hyper-parameter $K=32$. The values for $M$ and $\alpha$ will be discussed in Section \ref{sec2}.

\subsection{Bridging Transfer Learning}

FGVC tasks often face a domain gap: the general knowledge derived from large-scale, generalized datasets like ImageNet~\cite{deng2009imagenet} does not seamlessly transfer to the more specific and detailed knowledge required for smaller FGVC datasets \cite{wah2011caltech, krause20133d, maji2013fine}. To mitigate this issue, we propose a two-stage transfer learning strategy aimed at refining this domain difference. 
Initially, we use a pre-trained model ($\mathcal{M}_{pre}$) to
fine-tune a bridging model ($\mathcal{M}_{brg}$), followed by further refining the bridging model to obtain the final classification model ($\mathcal{M}_{cls}$). The training function $\tilde{\mathcal{M}}\leftarrow\Theta(\mathcal{M}, \mathcal{L})$ denotes the fine-tuning of model $\mathcal{M}$ on dataset $\mathcal{L}$ to obtain the updated model $\tilde{\mathcal{M}}$. The process is defined as follows:
\begin{equation}
	\begin{gathered}
		\mathcal{M}_{brg} \leftarrow \Theta(\mathcal{M}_{pre}, \mathcal{L}_{\alpha}),\\
		\mathcal{M}_{cls} \leftarrow \Theta(\mathcal{M}_{brg}, \mathcal{L}_{0}).
	\end{gathered}
\end{equation}

During the adaptation training phase, the model is trained with the augmented dataset $\mathcal{L}_{\alpha}$, which includes a mix of real and synthetic images at a rate defined by $\alpha$. In the final fine-tuning phase, the model is fine-tuned using a dataset $\mathcal{L}_{0}$ that contains only real images. This two-stage approach ensures that the model not only benefits from the variability introduced by the augmented data but also retains a strong alignment with the nuanced characteristics of the real-world FGVC datasets.

\section{Experiments}
\label{sec:Experiments}

This section describes our experiments in four key areas: (1) In Section \ref{sec2}, we examine the impact of the balance rate $\alpha$ of augmented image samples in the training dataset, both with and without our proposed Bridging Transfer Learning (BTL) process. (2) In Section \ref{sec2_5}, we control the external factors of the backbone model, base image augmentation, and input image size to evaluate the effectiveness of our proposed SGIA under different conditions and compare the performance with GIA. (3) Section \ref{sec3} investigates the integration of SGIA with large-scale CNN networks, comparing FGVC accuracies against image-based GIA and other state-of-the-art methods. (4) In Section \ref{sec4}, we analyze both positive and negative image samples generated, contrasting them with those produced by image-based GIA. To set the context, we first provide experiment details in Section \ref{sec1}.

\subsection{Dataset and Implementation Details}
\label{sec1}

Our experiments are conducted on three widely recognized FGVC datasets, including CUB-200-2011 Bird dataset \cite{wah2011caltech}, FGVC-Aircrafts \cite{maji2013fine}, and Stanford Cars \cite{krause20133d}. Each dataset comes with a predefined train-test split (except for few-shot evaluations in Section \ref{sec2}, in which we duplicate the corresponding split from \cite{azuri2021generative}). Unlike some previous works, e.g., \cite{hu2019better, diao2022metaformer}, which uses extra annotations like bounding box, segmentation, or meta information provided by these datasets, we use only the category labels for all model training.

Experiments are carried out on the Pytorch platform in Python. We utilize the pre-trained base stage model from I2VGen-XL\cite{zhang2023i2vgen} as our data augmentor. Unless otherwise specified, we employ random horizontal flipping and ``RandomResizedCrop'' (scale=(0.5, 1), imgsize=$224^2$) in Pytorch for training. In the testing phase, we resize the input image to have its shorter side be 256 pixels, and then center crops it to $224^2$. The training batch size is 16, with a weight decay of $1\times10^{-5}$. An initial learning rate of 0.01 is applied to all layers. We use a SGD optimizer and a cosine annealing scheduler with $t_0=1$ and $t_{multiply}=2$. The maximum epoch number is 128, with testing conducted at the end of each epoch. 


\subsection{Configuration and Comparison with the Baseline}
\label{sec2}

\begin{figure*}[t!]
	\centering
	\includesvg[width=\linewidth]{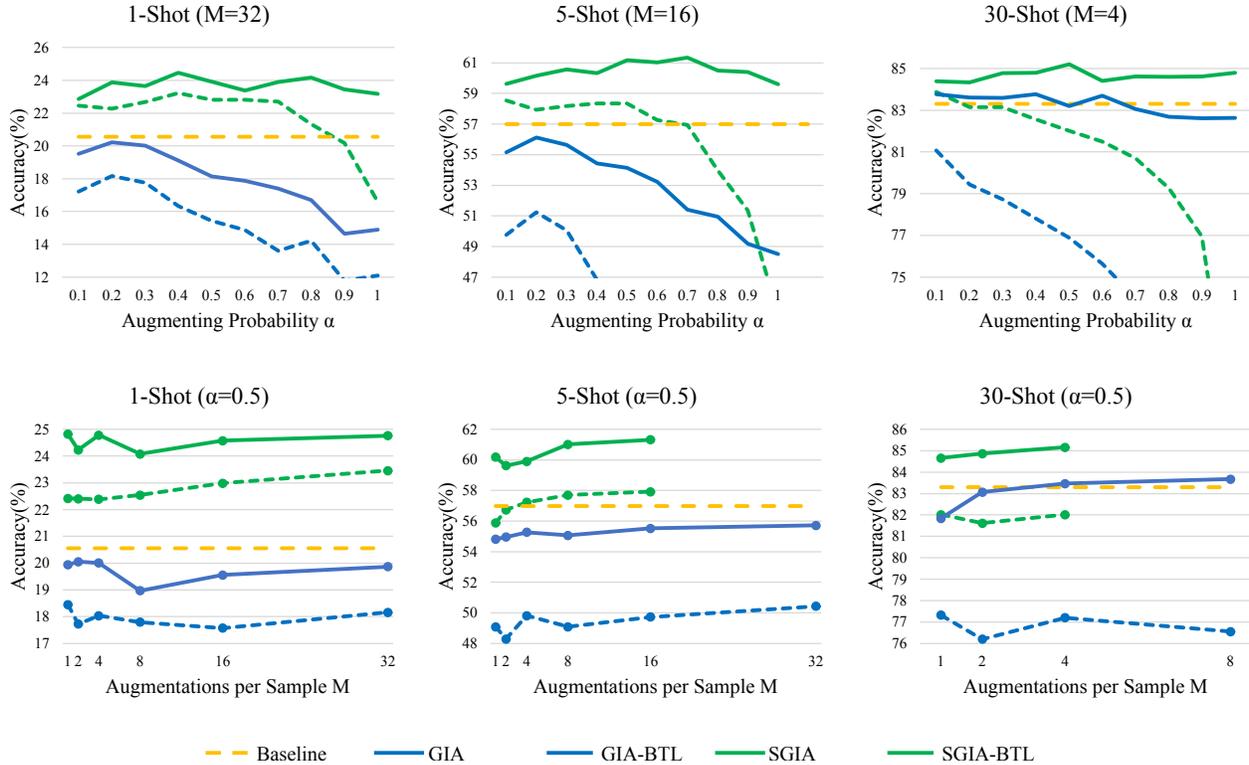}
	\caption{FGVC accuracies on CUB-200-2011 dataset \cite{wah2011caltech} of the proposed SGIA vs. different configurations of augmentation probability $\alpha$ and augmentations per sample $M$, and comparison with GIA (real guidance \cite{he2023is}).}
	\label{CONFIG_COMP}
\end{figure*}

The augmentation probability $\alpha$ and the number of augmentations per sample $M$, reflect the extent of inclusion of generative image samples in the training process and the count of augmented images created for each real image in the dataset, respectively. An increase in the $\alpha$ value enhances the model's variation and generalization capabilities during training but also leads to a greater presentation bias, as noted in \cite{he2023is}. A higher $M$ value results in improved variations but incurs additional computational cost linearly. To examine the effects of these variables, experiments were carried out varying $\alpha$ from 0.1 to 1.0 in increments of 0.1, and $M$ from 1 to 32 in doubling steps, on the CUB-200-2011\cite{wah2011caltech} dataset, using the EfficientNet-B0~\cite{tan2019efficientnet} model as a benchmark. Furthermore, the study compared the efficacy of the proposed SGIA against real guidance, a method of GIA referenced in \cite{he2023is}. For an equitable evaluation, all models were configured identically during training. The baselines are obtained with the maximum accuracy achieved by a single training phase and BTL. 

Results depicted in the first row of Fig. \ref{CONFIG_COMP} reveal that, compared to the benchmark, SGIA enhances the model precision at lower $\alpha$ values for few-shot and comprehensive FGVC datasets. Applying transfer learning via a bridging model yields accuracy enhancements of a maximum of +3.9\% at $\alpha=0.4$ for 1-shot, +4.4\% at $\alpha=0.7$ for 5-shot, and +1.9\% at $\alpha=0.5$ for the full dataset. Across the board, SGIA's performance surpasses that of GIA by up to 11.1\%. Consequently, $\alpha$ was set to 0.5 for subsequent experiments within this paper. The second row in Fig. \ref{CONFIG_COMP} illustrates the correlation between performance and augmentations per sample $M$. Accuracies for both SGIA and GIA generally escalate with $M$, yet SGIA with $M = 1$ exceeds both the baseline and GIA with $M = 32$ in every scenario. 

These experiments demonstrate SGIA's superiority over the baseline and GIA across all the FGVC scenarios evaluated. SGIA preserves the representational quality of the FGVC dataset while enhancing generalization capabilities through bridging transfer learning.

\subsection{External Variable Controls in Model Training}
\label{sec2_5}

\begin{table*}[t!]
	\centering
	\caption{Accuracies with controlled external variable in model training.}
	\begin{tabular}{ccc|ccc|ccc|ccc}
		\hline
		\textbf{BB.} & \textbf{Base} & \textbf{Img}& \multicolumn{3}{c|}{\textbf{CUB}} & \multicolumn{3}{c|}{\textbf{Aircrafts}} & \multicolumn{3}{c}{\textbf{Cars}} \\\cline{4-12} 
		\textbf{Model}	&\textbf{Aug.} &\textbf{Size} & \textbf{BaseL} & \textbf{GIA} & \textbf{SGIA} & \textbf{BaseL} & \textbf{GIA} & \textbf{SGIA} & \textbf{BaseL} & \textbf{GIA} & \textbf{SGIA} \\\hline\hline
		Res-18 & None & $224^2$ & 78.7 & 77.4 & \textbf{79.3} & 83.0 & 82.9 & 82.4 & 86.4 & 86.2 & \textbf{88.6} \\
		Res-18 & None & $448^2$ & 83.7 & 83.5 & \textbf{84.3} & 88.8 & 88.3 & \textbf{88.8} & 90.9 & 90.6 & \textbf{91.7} \\
		Res-18 & RRC & $224^2$ & 79.3 & 78.7 & \textbf{80.7} & 82.3 & 80.1 & \textbf{82.3} & 89.5 & 87.8 & \textbf{90.8} \\
		Res-18 & RRC & $448^2$ & 84.9 & 84.8 & \textbf{85.6} & 89.7 & 88.2 & \textbf{89.7} & 92.3 & 92.0 & \textbf{92.8} \\\hline
		Res-50 & None & $224^2$ & 81.9 & 81.6 & \textbf{82.5} & 84.6 & 83.9 & \textbf{84.8} & 88.4 & 87.5 & \textbf{90.3} \\
		Res-50 & None & $448^2$ & 86.3 & 85.3 & \textbf{86.3} & 90.5 & 89.6 & 89.9 & 92.0 & 91.9 & \textbf{93.0} \\
		Res-50 & RRC & $224^2$ & 82.8 & 81.8 & \textbf{83.1} & 85.8 & 84.8 & 85.4 & 91.5 & 89.5 & \textbf{91.7} \\
		Res-50 & RRC & $448^2$ & 86.7 & 86.6 & \textbf{87.1} & 91.1 & 90.6 & \textbf{91.6} & 93.2 & 93.0 & \textbf{93.5} \\\hline
		Eff-B0 & None & $224^2$ & 83.4 & 82.9 & \textbf{84.3} & 86.8 & 86.4 & \textbf{87.4} & 89.4 & 89.1 & \textbf{91.5} \\
		Eff-B0 & None & $448^2$ & 86.6 & 86.3 & \textbf{87.1} & 91.1 & 90.2 & \textbf{91.9} & 91.2 & 91.9 & \textbf{93.2} \\
		Eff-B0 & RRC & $224^2$ & 83.2 & 83.3 & \textbf{85.1} & 85.6 & 84.5 & \textbf{86.3} & 91.1 & 90.5 & \textbf{92.4} \\
		Eff-B0 & RRC & $448^2$ & 87.1 & 86.7 & \textbf{87.9} & 90.9 & 90.7 & \textbf{91.4} & 93.1 & 92.8 & \textbf{93.9} \\\hline
		Eff-B4 & None & $224^2$ & 84.5 & 84.6 & \textbf{86.1} & 88.8 & 88.9 & \textbf{88.9} & 89.7 & 89.1 & \textbf{91.5} \\
		Eff-B4 & None & $448^2$ & 88.3 & 87.9 & \textbf{88.5} & 91.8 & 92.3 & \textbf{92.2} & 91.8 & 92.2 & \textbf{93.4} \\
		Eff-B4 & RRC & $224^2$ & 85.6 & 84.6 & \textbf{85.7} & 87.8 & 87.5 & \textbf{87.8} & 91.3 & 90.3 & \textbf{92.1} \\
		Eff-B4 & RRC & $448^2$ & 88.5 & 88.3 & \textbf{88.5} & 91.7 & 91.7 & \textbf{92.3} & 93.2 & 93.3 & \textbf{93.9} \\\hline\hline
		
		\multicolumn{1}{c|}{}&\multicolumn{2}{c|}{Res-18} &---& -0.55 & \textbf{0.83} & --- & -1.08 & -0.15 & --- & -0.63 & \textbf{1.20} \\
		\multicolumn{1}{c|}{BB.}&\multicolumn{2}{c|}{Res-50} &---& -0.60 & \textbf{0.33} & --- & -0.83 & -0.08 & --- & -0.80 & \textbf{0.85} \\
		\multicolumn{1}{c|}{Model}&\multicolumn{2}{c|}{Eff-B0} &---& -0.28 & \textbf{1.03} & --- & -0.65 & \textbf{0.65} & --- & -0.13 & \textbf{1.55} \\
		\multicolumn{1}{c|}{}&\multicolumn{2}{c|}{Eff-B4} &---& -0.38 & \textbf{0.48} & --- & 0.05 & \textbf{0.25} & --- & -0.28 & \textbf{1.23} \\\hline
		
		\multicolumn{1}{c|}{Base}&\multicolumn{2}{c|}{None} &---& -0.49 & \textbf{0.63} & --- & -0.38 & \textbf{0.11} & --- & -0.16 & \textbf{0.68} \\
		\multicolumn{1}{c|}{Aug.}&\multicolumn{2}{c|}{RRC} &---& -0.41 & \textbf{0.70} & --- & -0.88 & \textbf{0.23} & --- & -0.75 & \textbf{0.74} \\\hline
		
		\multicolumn{1}{c|}{Img}&\multicolumn{2}{c|}{$224^2$} &---& -0.56 & \textbf{0.93} & --- & -0.73 & \textbf{0.08} & --- & -0.91 & \textbf{1.45} \\
		\multicolumn{1}{c|}{Size}&\multicolumn{2}{c|}{$448^2$} &---& -0.34 & \textbf{0.40} & --- & -0.53 & \textbf{0.26} & --- & 0.00 & \textbf{0.96} \\\hline
		
		\multicolumn{3}{c|}{Average Improvement} &---& -0.45 & \textbf{0.67} & --- & -0.63 & \textbf{0.17} & --- & -0.46 & \textbf{1.21} \\\hline
	\end{tabular}
	\label{CONTROL}
\end{table*}

In this section, we test our methods and compare them with the competitive GIA~\cite{he2023is} on various backbone models, base augmentations, input image sizes and FGVC datasets to illustrate the robustness of the proposed SGIA and BTL. We evaluate our model on two different CNN structures and two different network complexity for each structure: ResNet-18 (Res-18), ResNet-50 (Res-50)~\cite{he2016deep}, EfficientNet-B0 (Eff-B0) and EfficientNet-B4 (Eff-B4)~\cite{tan2019efficientnet}. We used various degrees of image augmentation as the basis for GIA and SGIA, where ``None'' refers to only using random horizontal flips, and ``RRC'' refers to using ``RandomResizedCrop'' (scale=(0.5, 1)) as the basic augmentation in addition to random horizontal flips. We utilized different input image sizes, i.e., $224^2$ and $448^2$, to verify the performance of SGIA under training data of different resolutions.

As shown in Table \ref{CONTROL}, our proposed SGIA surpasses or matches the baseline accuracy in 94\% of the experimental cases and exceeds the competitive method GIA\cite{he2023is} in 98\% of the cases, demonstrating excellent robustness across different datasets and training configurations. We calculate and display in Table \ref{CONTROL} the improvement levels of GIA and the proposed SGIA over the baseline under different controlled variables. The experimental results indicate that:

\begin{itemize}
	\item SGIA shows higher improvements for datasets with high deformations (e.g., +0.67\% for CUB-2011-200) and color variations (e.g., +1.21\% for Stanford Cars) compared to more rigid and less variable datasets (e.g., +0.17\% for FGVC-Aircraft). Benefiting from less representational variation, GIA is less impacted by the dataset features than SGIA.
	
	\item SGIA demonstrates greater enhancements for smaller scale networks (+0.62\% for ResNet-18 and +1.07\% for EfficientNet-B0) compared to larger networks (0.37\% for ResNet-50 and 0.65\% for EfficientNet-B4), which is due to the convergence of dataset accuracy. Simultaneously, we observed that for both GIA and SGIA, EfficientNet performs better than ResNet (0.87\% vs. 0.50\%), even though the selected EfficientNet models generally outperform ResNet on ImageNet-1K. We find that ResNet's larger number of parameters increases the risk of overfitting to augmented data, making SGIA more suitable for efficient networks with fewer parameters.
	
	\item Experiments show that SGIA performs better under stronger base augmentations (0.56 for RRC. vs. 0.48 for None), which is counterintuitive, as we usually consider that too strong augmentation might lead to underfitting risks and that the same extent of extra augmentation improves the model more with weaker base augmentation. However, given that the data added by GIA and SGIA are synthetic, a lower level of base augmentation might introduce systematic bias into the generative augmentation. For this reason, we argue that introducing a certain level of base augmentation when using SGIA could be more beneficial in enhancing model performance.
	
	\item Since the output image resolution of SGIA is $448\times256$, the enhancement to the model is weaker when the input size is $448^2$ compared to when the input size is $224^2$ (0.54\% vs. 0.82\%). However, even at a resolution of $448^2$, we still observe significant improvement in performance.
\end{itemize}

Our study demonstrates that the proposed SGIA method consistently outperforms the competitive GIA across various datasets, network architectures, and training configurations, proving its robustness and effectiveness in enhancing model performance with different levels of image augmentation and input sizes.

\begin{table*}[t!]
	\caption{Performance comparison on FGVC datasets. This table compares the classification performance of our proposed SGIA against baseline methods across various FGVC datasets. Results for previous works are replicated from their respective publications for comparative analysis.}
	\label{RSLT_full}
	\begin{center}
		\begin{tabular}{c|c c c|c c c }
			\hline
			\textbf{Method}&\textbf{Backbone}& \textbf{Pretrained} & \textbf{Size} & \textbf{CUB} & \textbf{Aircrafts} & \textbf{Cars}\\ \hline\hline
			WS-DAN \cite{hu2019better}  & Inception-v3\cite{szegedy2016rethinking} & ImageNet1k   &$448^2$    & 89.4  &  93.0  &  94.5     \\
			API-Net \cite{dong2020api}   &DenseNet-161\cite{huang2017densely} & ImageNet1k     &$448^2$     & 90.0   & 93.9  &  95.3   \\  
			AttNet \cite{hanselmann2020elope}  &ResNet-101\cite{he2016deep}& ImageNet1k   &$448^2$         & 88.9  &  94.1   & 95.6  \\  
			Mix+ \cite{li2020attribute})   &ResNet-50\cite{he2016deep}& ImageNet1k     &$448^2$     & 90.2   & 93.1   & 94.9  \\  
			TBMSL-Net \cite{zhang2020three}  &ResNet-50\cite{he2016deep}& ImageNet1k   &$448^2$        & 89.6   & 94.5   & 94.7  \\ 
			TransFG \cite{he2022transfg}   &ViT-B16\cite{dosovitskiy2020image} & ImageNet21k    &$448^2$     & 91.7   &  --- &   94.8        \\
			CAP \cite{behera2021context}  &Xception\cite{chollet2017xception}& ImageNet1k    &$224^2$     & 91.9    & 94.1  &   95.7        \\
			SR-GNN \cite{bera2022sr}   &ResNet-50\cite{he2016deep}& ImageNet1k  &$448^2$ & 91.9  & \textbf{95.4} &\textbf{96.1}\\ 
			MetaFormer \cite{diao2022metaformer}  &MetaFormer& iNat21\cite{inat2021}  &$384^2$  & 92.9   & 92.8  & 95.4   \\\hline
			Baseline\cite{woo2023convnext}   &ConvnextV2-H& ImageNet21k     &$512^2$      & 92.8    & 93.9      & 94.7     \\
			GIA(M=10) \cite{he2023is}  &ConvnextV2-H& ImageNet21k  &$512^2$      & 92.6    & 91.5      & 94.5 \\
			\textbf{SGIA}(M=3)  &ConvnextV2-H& ImageNet21k &$512^2$& \textbf{93.0}    & \textbf{94.1}      &\textbf{94.9}  \\
			\textbf{SGIA}(M=3)  &ConvnextV2-H& NABirds\cite{van2015building} &$512^2$ & \textbf{93.4}    &  ---    &  ---    \\\hline
		\end{tabular}
	\end{center}
\end{table*}

\begin{figure*}[t!]
	\centering
	\includesvg[width=\linewidth]{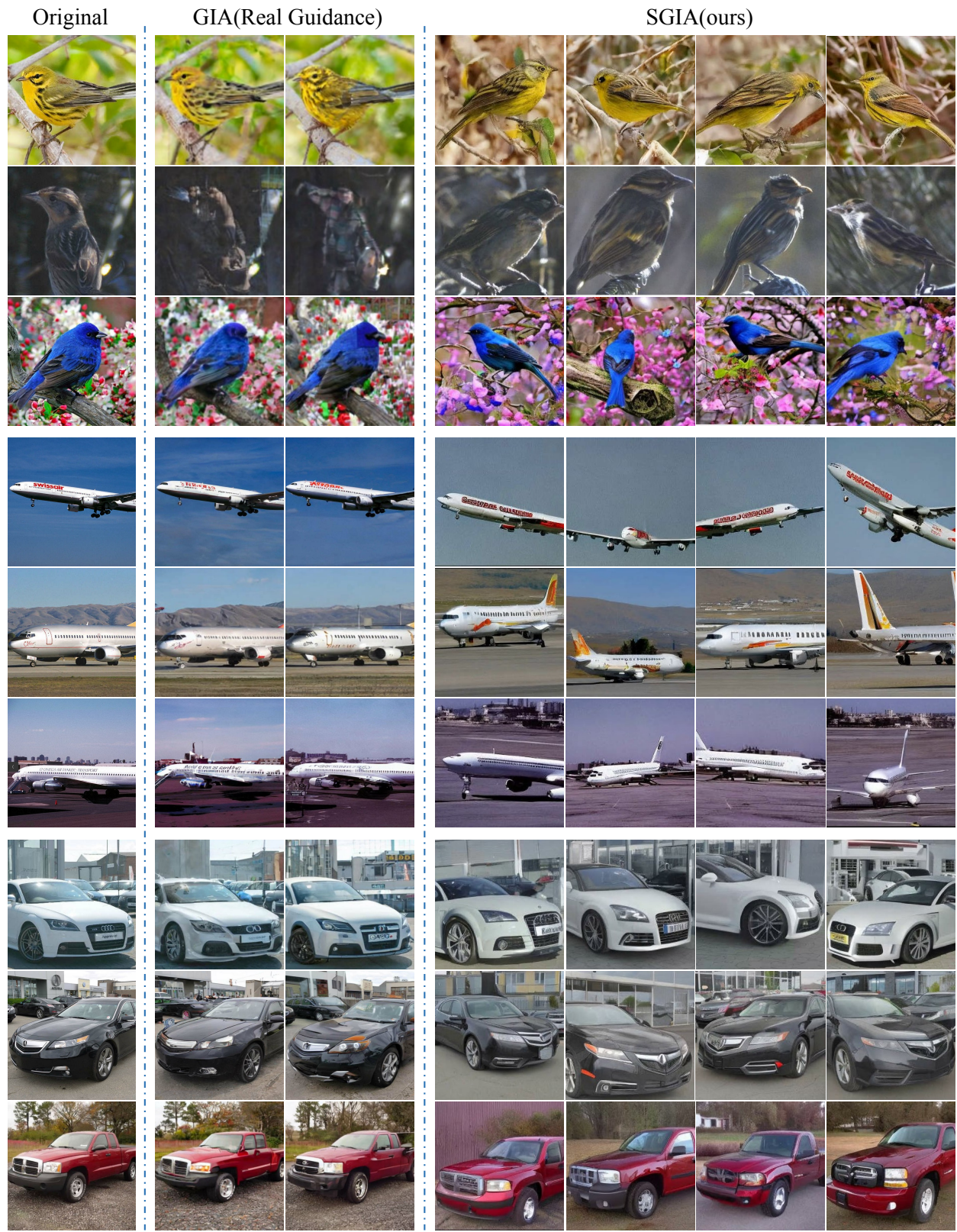}
	\caption{Generatied samples from GIA and SGIA. }
	\label{Positivce}
\end{figure*}

\subsection{SGIA for General FGVC}
\label{sec3}

In this section, we adopt the novel ConvnextV2-H\cite{woo2023convnext} model as our baseline to challenge the current state-of-the-art (SOTA) across multiple Fine-Grained Visual Categorization (FGVC) datasets, utilizing SGIA and contrasting with prior GIA. Our experimental approach largely adheres to the training protocols outlined in Section \ref{sec1}, with the notable adaptation of a two-step training strategy. Initially, we focus on training the fully connected classification head, subsequently progressing to fine-tune the entire network, applying the same hyperparameters. Specifically, for the SGIA (or GIA) and BTL methodologies, the first step involves employing SGIA (or GIA) to train the classification head, followed by comprehensive fine-tuning of the entire CNN network utilizing the original dataset. To accommodate larger model sizes and higher input resolutions ($512^2$), we adjust the batch size to 8 in this section.

The comparison of SGIA's performance against GIA and other FGVC models is presented in Table \ref{RSLT_full}. Against baselines set by ConvnextV2 \cite{woo2023convnext}, which are closely aligned with state-of-the-art models and with limited improvement scope, SGIA managed a 0.2\% boost in accuracy across three datasets, setting new records for the CUB-200-2011 dataset. This is in contrast to GIA, which, even with bridging transfer learning, diminished baseline accuracy. Further pretraining on the NABird dataset elevated CUB-200-2011 dataset accuracy to 93.4\%, surpassing the previous highest state-of-the-art accuracy by 0.5\%, as established by MetaFormer \cite{diao2022metaformer}. It's imperative to note that this leap in performance was attained with a substantially smaller pre-trained dataset and avoid using extra meta-annotations during training. 

\begin{figure}[t!]
	\centering
	\includesvg[width=\linewidth]{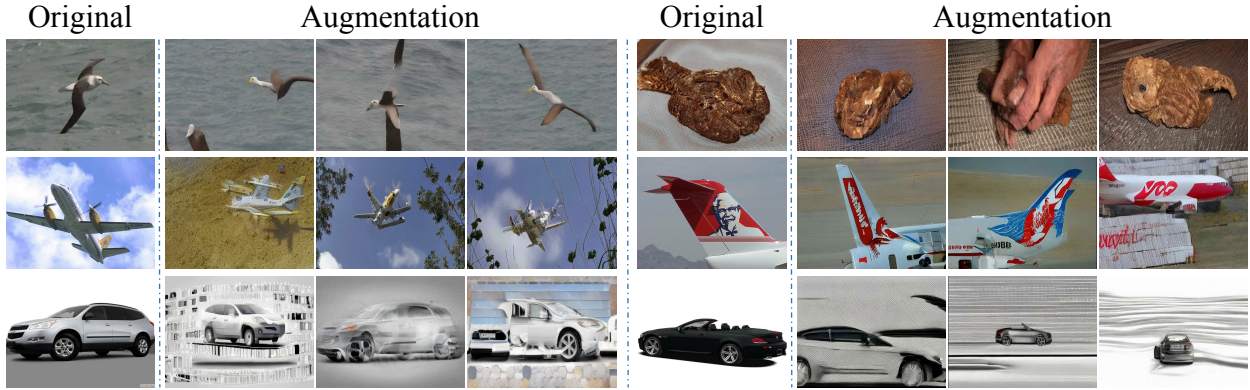}
	\caption{Negative samples from SGIA. The "Original" column displays real images from the three FGVC datasets. The "Augmentation" column shows negative samples generated by SGIA, characterized by less distinguishable features or lower image quality.}
	\label{Negative}
\end{figure}

\subsection{Image Augmentation Samples}
\label{sec4}

Three images from each of the three FGVC datasets mentioned in Section \ref{sec1} are randomly selected to produce augmentation samples using our SGIA and the image-based GIA from \cite{he2023is}, as illustrated in Fig. \ref{Positivce}. The left column shows three real images from each dataset. The middle columns present augmented samples generated by the method from \cite{he2023is} (two samples per real image), and the right columns feature samples generated by our proposed SGIA (four samples per real image), demonstrating the enhanced diversity and realism from SGIA. Compared to the original images, augmented samples from Real Guidance \cite{he2023is} maintain the primary composition, introducing minor variations in texture and background. In contrast, SGIA introduces more extensive variations, including changes in viewpoint, position, action, lighting, and even the shape of interacting objects (e.g., branches under the bird's feet). Additionally, SGIA samples exhibit clearer and more natural presentations, suggesting a narrower gap between augmented and real images, contributing to improved FGVC accuracies.

Negative samples generated by SGIA are depicted in Fig. \ref{Negative}. These include instances where the major feature is missing, and indistinct or irrelevant features appear in the augmented images. Such negative generation mainly comes from the lack of spatiotemporal consistency from the generative model and can impact the representational capability of models trained on augmented datasets. It's noteworthy that this issue of information loss is not unique to SGIA but is also common in image-based GIA and traditional augmentations like random erasing \cite{zhong2020random}.
\section{Conclusion}
\label{sec:Conclusion}

This paper introduces Sequence Generative Image Augmentation (SGIA), a novel method for Fine-Grained Visual Categorization that diversifies perspectives, backgrounds, and object interactions while preserving key features. Leveraging the Bridging Transfer Learning (BTL) framework, we effectively mitigate the influence of systemic data distribution biases inherent in SGIA, thereby bolstering the generalizability of models trained with this method. Our methodical experimentation, using a controlled variable approach, assesses SGIA's effectiveness in bolstering baseline models across diverse datasets, model architectures, augmentation extents, and training parameters, affirming its adaptability to a wide range of external conditions. Comparative analyses reveal that SGIA outclasses traditional image-based Generative Image Augmentation (GIA) strategies in generating high-quality and diverse images, making FGVC models more robust to real-world variations. SGIA consistently exceeds conventional methods under both few-shot and comprehensive data scenarios, setting a new benchmark in the CUB-200-2011 dataset and advancing the field of image augmentation for FGVC tasks.

\vfill\pagebreak

\bibliographystyle{unsrt}  
\bibliography{ref}

\end{document}